\documentclass[11pt,a4paper]{article}
\usepackage[hyperref]{acl2020}
\usepackage{times}
\usepackage{latexsym}
\usepackage{multirow}
\usepackage{graphics}
\usepackage{graphicx,subfigure}
\usepackage{microtype}

\aclfinalcopy % Uncomment this line for the final submission
\title{Transformer on a Diet}

\author{
 Chenguang Wang \qquad Zihao Ye \qquad Aston Zhang \\ {\bf Zheng Zhang \qquad Alexander J. Smola} \\
  Amazon Web Services \\
  \texttt{\{chgwang, yeziha, astonz, zhaz, smola\}@amazon.com} \\
}
\date{}

\begin{document}

\maketitle

\begin{abstract}
Transformer has been widely used thanks to its ability to capture sequence information in an efficient way. However, recent developments, such as BERT and GPT-2, deliver only heavy architectures with a focus on effectiveness. In this paper, we explore three carefully-designed light Transformer architectures to figure out whether the Transformer with less computations could produce competitive results. Experimental results on language model benchmark datasets hint that such trade-off is promising, and the light Transformer reduces 70\% parameters at best, while obtains competitive perplexity compared to standard Transformer. The source code is publicly available~\footnote{\url{https://tinyurl.com/qkp3zf3}}.
\end{abstract}
\section{Introduction}

Transformer has shown its effectiveness in modeling sequence information due to the combination of self-attention mechanism and positional encoding. The variants of Transformer architecture, such as BERT \cite{devlin2018bert} and GPT-2 \cite{radford2018improving,radford2019language}, have obtained the state-of-the-art results across a wide range of NLP tasks, including GLUE benchmark dataset \cite{wang2018glue}, and question answering dataset, e.g., SQuAD \cite{rajpurkar2016squad}. However, Transformer in nature is a fully-connected feed-forward neural network and exhibits heavy computation characteristics. The recent BERT and GPT-2 are constructed as a stack of Transformer blocks, e.g., the largest GPT-2 is a stack of 48 Transformer blocks and contains 1542M parameters, BERT-large contains 24 layers of Transformer block and results in 340M parameters, therefore the computational burden of the fully connected Transformer becomes heavier. A side effect in industrial applications is that this potentially makes it harder to deploy due to the huge size of the model. Therefore a light version of the standard Transformer architecture is expected to relieve the heavy computation issue and compress the model to ease the deployment in real world applications.

In this paper, we carefully design several light Transformer architectures. The intuition behind the light Transformers is: preserving the Transformer connections that are useful to capture the essential sequence information, while omitting the ones with less impact. In particular, we explore along two directions: 1) better preserve the connections that are useful for capturing long-range dependency. We adapt the idea of dilated convolutions \cite{yu2015multi} to preserve the Transformer connections that are useful to extend the effective history of the context, and 2) better preserve the connections that are essential in capturing local context. We leverage cascade connections that are capable to intensively incorporate the local context information in a flexible manner. 

The contributions of the paper are two-folds:
\begin{itemize}
    \item We explore three light Transformer architectures that are able to preserve the necessary connections in standard Transformer. We show that the light Transformer architectures reduce the computation from quadratic to linear compared to the standard Transformer.
    \item We conduct experiments on two language model benchmark datasets, one of the most traditional sequence modeling tasks, where the results indicate that the lightest architecture could reduce 70\% parameters of standard Transformer, and performs competitively with the standard Transformer.
\end{itemize}

% We also exploit the design that could jointly preserve the long-range and local dependency, where we split the sequence into different granularity represented as a hierarchical tree, then aggregate the context at different granularity.

\section{Light Transformers}

We describe the three proposed light Transformer architectures in this section.

\subsection{Background}

{\bf Revisiting Transformer architecture.} Transformer \cite{vaswani2017attention} consists of an encoder and a decoder. We mainly focus on the sequence generation problem, thus we briefly describe the Transformer decoder structure, full Transformer, below for sake of clarity, in the following sections, we mean Transformer decoder when mention Transformer, unless other ways stated. As illustrated in Figure~\ref{fig:fulltrans}, the full Transformer block contains two sub-layers: 1) a masked multi-head attention layer; 2) a position-wise fully connected feed-forward network. Besides, there is a residual connection \cite{he2016deep} around each of the two sub-layers, followed by layer normalization \cite{ba2016layer}. 

{\bf Transformer computation complexity analysis.} For each Transformer block, we assume that the length of the sequence to be $n$, the size of hidden states to be $h$, the computation of each Transformer block is $O(n^2 \cdot h)$.

We regard the following three architectures that have less computation compared to the full Transformer as light Transformers.

\subsection{Dilated Transformer}

{\bf Model assumption.} The key strength of the Transformer is that the combination of self-attention and positional encoding is able to capture long-term dependency in the sequence. Thus we need to preserve the long-term dependency when lightening up the Transformer blocks. Inspired by the dilated convolutions \cite{yu2015multi}, we introduce the idea of dilated Transformer to enable an exponentially large receptive field in the Transformer scenario. 

{\bf Model architecture.} The model architecture is illustrated in Figure~\ref{fig:lighttransa}, where the sub-layers are the same to full Transformer, but with dilated connections across the sequence. To enable this, we introduce $d$ as the dilation factor, $k$ as the filter size. Similar to the common usage in dilated convolutions, we increase $d$ exponentially with the depth of the Transformer based network, i.e., $d = O(2^l)$ at level $l$ of the Transformer, to increase the receptive field. By doing this, there is some filter that hits each input within the effective history, while also allowing for an extremely large effective history using the deep Transformer architecture.

{\bf Computation complexity analysis.} In each dilated Transformer block, there would be $k$ nodes need to compute the output of the current node, so the computation complexity is $O(n\cdot k \cdot h)$. The computation cost is significantly lower compared to that of the full Transformer when $k$ is significantly smaller than $n$.

\subsection{Dilated Transformer with Memory}

{\bf Model assumption.} Similar to the idea of dilated Transformer, we use dilated connections to preserve the long-range dependency in the sequence. Additionally, in dilated Transformer with memory, we try to cache more local contexts by memorizing the nodes in the previous dilated connections.

\begin{figure*}[]
     \begin{center}
        \subfigure[Full Transformer.]{%
            \includegraphics[scale=0.4]{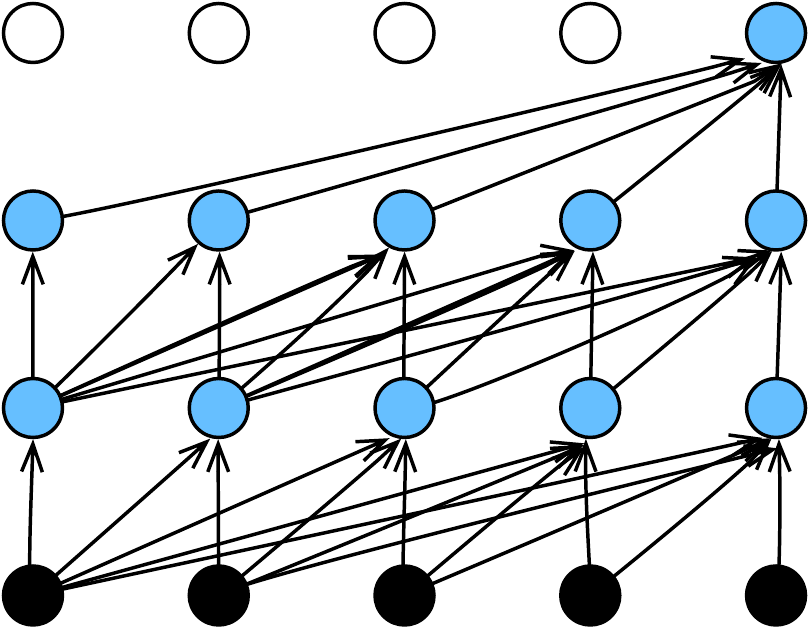}
            \label{fig:fulltrans}
        }%
        \hspace{0.1in}
        \subfigure[Dilated Transformer.]{%
            \includegraphics[scale=0.4]{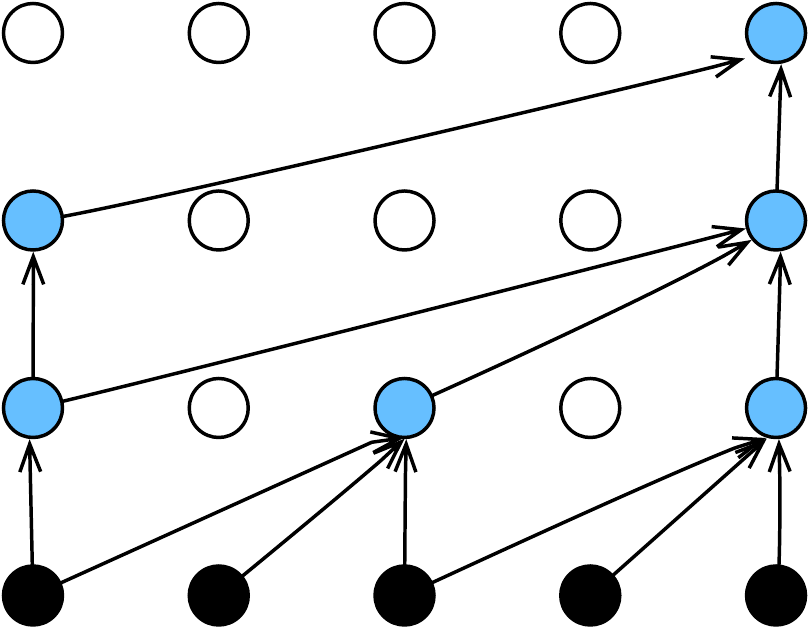}
            \label{fig:lighttransa}
        }%
        \hspace{0.1in}
        \subfigure[Dilated Transformer with memory.]{%
            \includegraphics[scale=0.4]{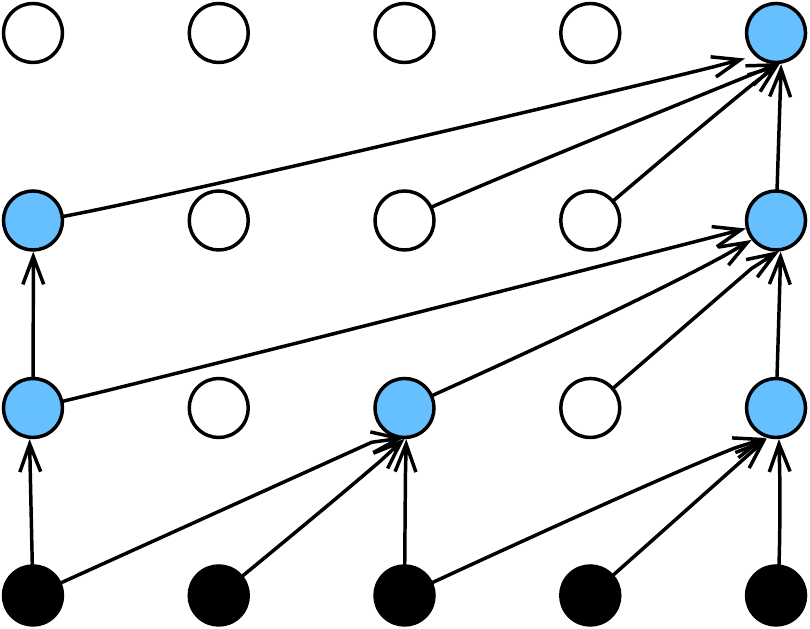}
            \label{fig:lighttransb}
        }%
        \hspace{0.1in}
        \subfigure[Cascade Transformer.]{%
           \includegraphics[scale=0.4]{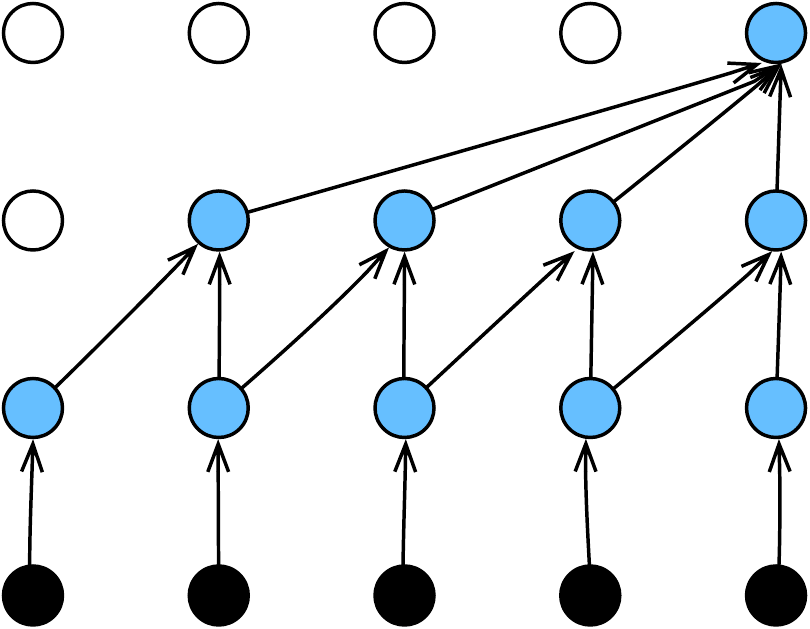}
           \label{fig:lighttransc}
        }%
        % \hspace{0.2in}
        % \subfigure[SegTree Transformer.]{%
        %     \includegraphics[scale=0.5]{figures/aggregated-tf.pdf}}
        % }%
    \end{center}
    \caption{%
        Transformer architectures. (a): standard Transformer; (b)-(d): proposed light Transformers.
     }%
   \label{fig:lighttrans}
\end{figure*}

{\bf Model architecture.} Figure~\ref{fig:lighttransb} illustrates the model architecture, where the sub-layers are still the same with full Transformer. Similar to dilated Transformer, we use the dilation factor and filter size to construct the dilated connections. However, the dilated connections in the previous layer are preserved. This will ensure a large effective history by with richer local history. This could potentially preserve the connections that are necessary to decode.

{\bf Computation complexity analysis.} In each dilated Transformer with memory block, the computation of the connections in the previous layer would add to the current computation, which results in $O(n\cdot k \cdot c \cdot h)$, where $c$ indicates the extra connections in the previous layer. If the sequence length is infinite, then $c = 2$.

% \begin{figure}[]
%     \centering
%     \includegraphics[scale=0.5]{figures/tf.pdf}
%     \caption{Full Transformer. 
%     \label{fig:fulltrans}}
% \end{figure}

\subsection{Cascade Transformer}

{\bf Model assumption.} Instead of exploiting the dilated Transformer idea, we instead explore cascade connections idea to exponentially incorporate the local connections. By exploring this method, we would see how the local connections in different depths of the network contribute to the results.

{\bf Model architecture.} Figure~\ref{fig:lighttransc} illustrate the cascade Transformer architecture, where the sub-layers are still the same as full Transformer. We introduce base window size as $b$, the cardinal number is $m$, then the number of cascade connections at level $l$ of the Transformer is $O(b \cdot m^l)$. By doing this, we can control the shape of the cascade across the levels of the Transformer, which gives Transformer the flexibility to learn from the cascade connections.

{\bf Computation complexity analysis.} In each cascade Transformer, the computation cost is $O(n \cdot b \cdot m^l \cdot h)$. Compare to full Transformer, the complexity is still smaller since $b \cdot m^l < n$.

\begin{table}[]
\centering
\small
\begin{tabular}{|l|l|}
\hline
\textbf{Model}      & \textbf{Computation Complexity} \\ \hline
Full                &  $O(n^2 \cdot h)$                               \\ \hline
Dilated             &  $O(n\cdot k \cdot h)$                               \\ \hline
Dilated-Memory      &  $O(n\cdot k \cdot c \cdot h)$                               \\ \hline
Cascade             &  $O(n \cdot b \cdot m^l \cdot h)$                               \\ \hline
% SegTree Transformer & $O(n \cdot \log n \cdot h)$                                \\ \hline
\end{tabular}
\caption{Computation complexities of different Transformer architectures. Full: full Transformer; Dilated: dilated Transformer; Dilated-Memory: dilated Transformer with memory; Cascade: cascade Transformer; $n$ is the length of the sequence. $h$ is the size of the hidden state. $k$ is the filter size. $b$ is the base window size. $m$ is the cardinal number. }
\end{table}
% SegTree: SegTree Transformer. 

\section{Transformer Language Model} 

We select language model as the task to evaluate the proposed Transformer architecture, since it is one of the fundamental NLP tasks. In this section, we introduce how the different Transformer blocks adapted to the task of language model.

Given a corpus of tokens $\mathbf{X} = (X_1, \dots, X_T)$, the objective of language model is described in Eq.~\ref{eq:lm}.
\begin{equation}
    \mathcal{L}(\mathbf{X}) = \sum_t log P(X_t | X_{<t})
    \label{eq:lm}
\end{equation}

Then the Transformer (decoder) blocks are used to generate the output distribution over the vocabulary as indicated in Eq.~\ref{eq:trans}.
\begin{equation}
    h_0 = \mathbf{X} W_e + W_p
\end{equation}
\begin{equation}
    h_l = {\rm transformer\_block}(h_{l-1})
\end{equation}
\begin{equation}
    p(X) = {\rm softmax}(h_L {W_e}^T)
    \label{eq:trans}
\end{equation}
where ${\rm transformer\_block}$ can be replaced with any of the three proposed Transformer architectures, $h_l$ is the hidden output of $l$-th layer, $W_e$ is the word embedding matrix, and $W_p$ is positional embedding matrix.

\section{Experiments}

\begin{table*}[]
\centering
\begin{tabular}{|l|l|l|l|l|l|}
\hline
\multicolumn{1}{|c|}{\multirow{2}{*}{\textbf{Model}}} & \multirow{2}{*}{\textbf{Parameter}} & \multicolumn{2}{c|}{\textbf{PTB}}                                      & \multicolumn{2}{c|}{\textbf{WT-2}}                                     \\ \cline{3-6} 
\multicolumn{1}{|c|}{}                                &                                     & \multicolumn{1}{c|}{\textbf{Val}} & \multicolumn{1}{c|}{\textbf{Test}} & \multicolumn{1}{c|}{\textbf{Val}} & \multicolumn{1}{c|}{\textbf{Test}} \\ \hline
Full                                                  & 30.0M                               & 109.19                            & {\bf 103.72}                             &  148.76                                 &     140.74                               \\ \hline
Dilated                                               & {\bf 8.8M}                                & 115.67                            & 110.92                             &  157.67                                 &   147.58                                 \\ \hline
Dilated-Memory                                        & 11.1M                               & 115.35                            & 110.98                             &      167.35                             &       157.08                             \\ \hline
Cascade                                               & 13.5M                               & 109.16                            & 105.27                             &     145.96                              &       {\bf 136.02}                             \\ \hline
% SegTree                                   & 27.0M                               & {\bf 96.84}                             & {\bf 90.86}                              & {\bf 113.32}                            & {\bf 107.69}                             \\ \hline
\end{tabular}
\caption{Results comparison (perplexity) of different Transformer language models on PTB and WT-2 data. Full: full Transformer; Dilated: dilated Transformer; Dilated-Memory: dilated Transformer with memory; Cascade: cascade Transformer.}
\label{tab:result}
\end{table*}

We compare the light Transformers with standard Transformers from both results and computation perspectives.

\subsection{Datasets and Metrics}
We evaluate the proposed methods on three widely-used language model
benchmark datasets. {\bf Penn TreeBank (PTB)}: we use the preprocessed
version of \cite{mikolov2010recurrent}, which contains 100M
tokens. {\bf WikiText-2 (WT-2)} is a small preprocessed version of
Wikipedia, containing 200M tokens \cite{MerityXBS16}.
We use perplexity to evaluate the language model results.

% \begin{figure*}[ht!]
%      \begin{center}
%         \subfigure[PTB.]{%
%             \includegraphics[scale=0.6]{PTB}
%         }
%         \hspace{0.1in}
%         \subfigure[WT-2.]{%
%             \includegraphics[scale=0.6]{WT2}
%         }
%     \end{center}
%     \caption{%
%         Results and parameter size correlation of different Transformer architectures. Full: full Transformer; Dilated: dilated Transformer; Dilated-Memory: dilated Transformer with memory; Cascade: cascade Transformer; SegTree: SegTree Transformer.
%      }%
%   \label{fig:result}
% \end{figure*}

\subsection{Training Details}

For fair comparison, the full Transformer and the light Transformer architectures are with 3 layers, embedding size equals to 320, number of heads in the multi-head attention is 16. The dropout rate is set as 0.4 and 0.2 on PTB and WT-2 respectively. For dilated Transformer and dilated Transformer with memory, $k = 3$, the base of $d = 2$. For cascade Transformer, $b = 4$ and $m = 2$. For the light Transformers and full Transformer, the hidden size equals to 2000. These settings are shared on the two datasets. 

We use truncated back-propagation through time to compute the gradients across all the experiment settings. The batch size equals to 20 on both datasets, whereas the sequence length equals to 70 on both datasets. We use SGD for training with learning rate equals to 10.

%  other than SegTree. For SegTree Transformer, we use Adam with initial learning $10^{-3}$ as our optimizer, and divide the learning rate by half once loss on validation set does not diverge further. The aggregate function we select is tree-lstm cell

\subsection{Results Analysis}

We compare the effectiveness of the proposed Transformer architectures with the full-Transformer architecture. From the results in Table~\ref{tab:result}, we find out that cascade Transformer performs closely to the full-Transformer structure. This indicates that local context is very important to language model tasks, and cascade Transformer is able to capture the meaning local dependency. 

% SegTree Transformer consistently performs the best on both PTB and WT-2 datasets. Especially, SegTree Transformer outperforms full-Transformer. The result indicates that by encoding the inductive bias of attending more on effective neighbors and less on too far away context, the light Transformer is more robust to the sequence data compared to the fully-connected architecture.
% Notice that

We also compare the parameter sizes between light Transformers and full Transformer. Among all the architectures, dilated Transformer is lightest one. Although it delivers moderate results, however, when compare to full Transformer, we save 70\% model size and the computation could be more efficient. 
% Note that SegTree Transformer saves 10\% model size but gain on average 22.9 perplexities compared to full Transformer. 
Table~\ref{tab:result} also shows the trade-off between parameter size and perplexities on the two datasets. It would suggest the best Transformer architecture given the deployment constraints, such as model size limit, latency requirement or the quality.

\vspace{-0.05in}
\section{Related Work}

{\bf Transformer architectures} have been proposed to compute the sequence input efficiently.
The basic Transformer block consists of a multi-head attention layer and a position-wise fully connected feed-forward network. The original Transformer architectures contains an encoder and decoder. The encoder and decoder share similar structures with 6 layers of Transformer block. Instead, the decoder uses masked multi-head attention each block to prevent leftward information flow. Recently, stacked Transformer architectures, such as BERT \cite{devlin2018bert}, GPT(-2) \cite{radford2018improving,radford2019language}, and the most recent ones \cite{PetersNIGCLZ18,wang190409408,abs-1910-10683,abs-1907-11692,abs-1906-08237} are proposed and shown the state-of-the-art results on a wide range of NLP tasks, such as GLUE benchmark \cite{wang2018glue} and question answering datasets \cite{rajpurkar2016squad}. However, these Transformer architectures are heavy and it is hard to deploy in practice where the environment has constraints. Lightened Transformer architectures \cite{ye2019bp,guo2019star} are proposed to speed up the computation. Our work is aligned with such Transformers but with even less computation. Compared to ALBERT \cite{lan2019albert}, the proposed method optimizes the base Transformer and could be further integrated into BERT.
% TODO: zihao, if you have time, please also help on fill out this related work on Transformer.

{\bf Language models} have been studied extensively in NLP. Neural language models have supplanted traditional n-gram models in recent years \cite{bengio2003neural,mnih2007three,mikolov2010recurrent}. Particularly, recurrent neural networks \cite{inan2016tying,Merity02182,melis2017state,KrauseK0R18}, such as LSTMs have achieved state-of-the-art results on various benchmark datasets with different regularization techniques and post-training methods \cite{GraveJU16,KrauseK0R18}. 
The mixture of softmax \cite{Yang03953} has helped address the low-rank embedding problem for word prediction. Recently, more advanced Transformer architectures, such as GPT \cite{radford2018improving} and GPT-2 \cite{radford2019language} are applied to the task of language model. Due to the efficiency of the Transformer computation, these models have been trained on large scale text corpora and shown good results across language model datasets. We instead study how to lighten the Transformer, which could be generalized to the idea of large Transformer architecture (e.g., GPT) to train on large corpora to obtain better results.
\vspace{-0.05in}
\section{Conclusion}
\vspace{-0.05in}
We explore less computation-expensive Transformer architectures. The design principle is to still preserve the long and short range dependency in the sequence but with less connections. Experiments on language model datasets show that a light weighted Transformer is able to perform competitively but with much improved computation efficiency. We plan to extend the Transformer architectures to experiment on deeper Transformer architectures and more tasks~\cite{WangSERZH15,WangSLZH16}.

{\bf Acknowledgements} We are grateful to Mu Li, Da Zheng, Haibin Lin, and Leyuan Wang for their helpful inputs on the paper.

\bibliographystyle{acl_natbib}
\bibliography{acl2020}
\end{document}